\documentclass[letterpaper, 10 pt, conference]{ieeeconf}  

\IEEEoverridecommandlockouts  
\usepackage{graphics} 
\usepackage{epsfig} 
\usepackage{mathptmx} 
\usepackage{times} 
\usepackage{cite}
\usepackage{scrextend}

\usepackage{hyperref}
\usepackage{silence}
\usepackage{amsmath}
\usepackage{amssymb}
\usepackage{array}
\usepackage[warn]{textcomp}

\usepackage{amsthm}

\usepackage{centernot}
\usepackage[linesnumbered,ruled]{algorithm2e}
\SetKwRepeat{Do}{do}{while}
\SetKwComment{Comment}{$\#$}{}
\usepackage{graphicx}
\graphicspath{{images/}{graphs/}{graphs/charts/}{qmap/}}
\usepackage{wrapfig}
\usepackage{floatrow}
\usepackage{caption}
\usepackage{subcaption}
\usepackage{sidecap}
\usepackage{adjustbox}

\usepackage{booktabs}

\usepackage{listings}

\usepackage{color}
\usepackage[dvipsnames]{xcolor}

\definecolor{mygreen}{rgb}{0,0.6,0}
\definecolor{mygray}{rgb}{0.5,0.5,0.5}
\definecolor{mymauve}{rgb}{0.58,0,0.82}

\title{\LARGE \bf
Recurrent Deterministic Policy Gradient Method for Bipedal Locomotion on Rough Terrain Challenge
}

\author{ \parbox{6 in}{\centering Doo Re Song, Chuanyu Yang, Christopher McGreavy, and Zhibin Li
         \thanks{* Presented in \href{https://ieeexplore.ieee.org/document/8581309}{IEEE-ICARCV 2018}, Singapore: \tiny{DOI 10.1109/ICARCV.2018.8581309}}
         \\
		 School of Informatics, University of Edinburgh, UK\\
		 {\tt\small sdr2002@gmail.com, (chuanyu.yang, c.mcgreavy, zhibin.li)@ed.ac.uk}}
}

\begin{document}

\maketitle
\thispagestyle{empty}
\pagestyle{empty}

\begin{abstract}
This paper presents a deep learning framework that is capable of solving partially observable locomotion tasks based on our novel interpretation of Recurrent Deterministic Policy Gradient (RDPG). We study on bias of sampled error measure and its variance induced by the partial observability of environment and subtrajectory sampling, respectively. Three major improvements are introduced in our RDPG based learning framework: tail-step bootstrap of temporal difference, initialisation of hidden state using past subtrajectory, truncation of temporal backpropagation, and injection of external experiences learned by other agents. The proposed learning framework was implemented to solve the Bipedal-Walker challenge in OpenAI's gym simulation environment where only partial state information is available. Our simulation study shows that the autonomous behaviors generated by the RDPG agent are highly adaptive to a variety of obstacles and enables the agent to effectively traverse rugged terrains for long distance with higher success rate than leading contenders.
\end{abstract}

\section{Introduction}

The morphology of humanoid robots is similar to that of humans which provides the ability to traverse complex and dynamic terrains that are easily accessible to humans. Humanoid robots generally exhibit high manoeuvrability and flexibility, thus are capable of achieving locomotion while navigating through uneven terrain and stepping over obstacles. Considering the physical limitations of wheeled robots, there are many advantages to choosing bipedal locomotion over wheeled locomotion. Moreover, knowledge of bipedal locomotion can also help us design better exoskeletons that can benefit the lives of people with gait abnormalities. As a result, bipedal locomotion has attracted increasing attention in recent years.

Most bipedal walking controls are achieved using deterministic and analytic engineering approaches. Yet, there are also a large amount of works that attempt to use machine learning approaches, such as reinforcement learning (RL), to achieve bipedal walking. RL can be applied to model-free learning for bipedal walking. This has been attempted in various action policy learning algorithms based on Markov Decision Process (MDP) \cite{tedrake2004stochastic,morimoto2004simple}.

Increasing amounts of researchers are interested in investigating the potential of implementing Deep Reinforcement Learning (DRL) methods to allow agents to perform dynamic motor tasks in complex environments. DRL is a RL framework that combines reinforcement learning with deep neural networks. Within DRL, deep neural networks are utilized to approximate value function $V(s)$ or $Q(s,a)$, action policy $\pi(a|s)$, and state transition and reward model \cite{sutton2000policy}. DRL research has an growing interest in recent years due to its capability as a non-linear state abstraction tool which deals with action decision as well as evaluation. A great breakthrough in the development of DRL is its first human-level performance on simple Atari games by Mnih et al. \cite{mnih2015human}. As a proceeding, our motivation is at investigating the potential of implementing DRL methods to allow agents to perform dynamic motor tasks in complex environments, such as robot locomotion. 

A variety of work by computer science and robotics researchers has used DRL to solve bipedal locomotion. Peng et al. has successfully trained a bipedal humanoid character to traverse rugged terrain in a 2D simulation environment using Continuous Actor Critic Learning Automation (CACLA) \cite{peng2015dynamic}. They later extended their work to include a hierarchical DRL framework and trained a bipedal humanoid character to perform complex locomotion tasks in a 3D simulation environment \cite{peng2017deeploco} \cite{yang2017emergence}.

\begin{figure}[t]   
	\centering
	\begin{subfigure}[t]{0.3\textwidth}
		\includegraphics[width=\textwidth]{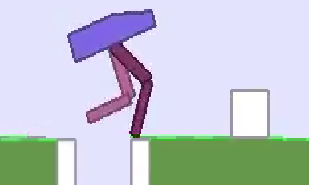}
		\caption{}
		\label{fig:poenv3}
	\end{subfigure}
	~ 	
	\begin{subfigure}[t]{0.26\textwidth}
		\includegraphics[width=\textwidth]{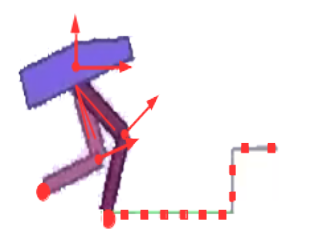}
		\caption{}
		\label{fig:poenv1}
	\end{subfigure}			
	~			
	\begin{subfigure}[t]{0.3\textwidth}
		\includegraphics[width=\textwidth]{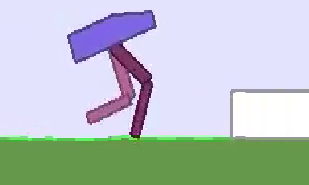}        
		\caption{}
		\label{fig:poenv2}
	\end{subfigure}

	\caption{Partially observable scenario: (a) Real world environment; (b) Incomplete observation of environment; (c) Incorrect perception built on the incomplete observation;.}\label{fig:poe}  
\end{figure}

However, the MDP framework of RL lacks some necessary components for walking. During sensing, a walking agent is unable to fully observe its environment (Fig.\ref{fig:poenv1}), meaning inferences must be made about the environment using previous observations and actions made by the agent to fill in the gaps in observability; the estimated state is then used to infer actions (Fig.\ref{fig:poenv3}), otherwise the agent can only act the same as it does in less risky circumstance (Fig.\ref{fig:poenv2}). However, the MDP process has no functionality to distinguish them. State of the art RL algorithms with MDP is able to succeed in fully observable virtual environments \cite{mnih2015human} but not in the partially observable environment. In reality however, robots never gather sufficient information to generate optimal actions, and generally the obtained information is usually noisy as well. Therefore, partial observability is a critical issue in robot locomotion when RL is applied to real world robotics. Advancements in the improved decision framework of agents allow reasonable estimation of the true state of the environment when full observability is not possible.


There are related works that introduced Partially Observable Markov Decision Process (POMDP)\cite{smallwood1973optimal} to RL to address the issue of incomplete observability in the environment. Wierstra et al. suggested an early form of Recurrent Neural Network (RNN) based RL, where the benefits of the POMDP framework provided by the standard RNN were reported in the partially observable environment though tasks were simple\cite{wierstra2009recurrent}. 
Mnih et al. reported that the RNN-based RL algorithms with a sensible observation-action evaluator can improve the learning performance of RL agents beyond human expert level if the observation is well provided\cite{mnih2016asynchronous}. There has been works that combined RNN with DRL to solve bipedal locomotion. Heess et al. proposed a RNN-based DRL called Recurrent Deterministic Policy Gradient (RDPG) \cite{heess2015memory} that is a combination of Long Short Term Memory network (LSTM) \cite{hochreiter1997long} and Deterministic Policy Gradient (DPG) \cite{silver2014deterministic}, and the emergence of locomotion behaviour was obtained on flat terrain environment. 

Finally, recent works applied RNN-based RL to locomotion learning tasks and observed emergence of walking behaviour from scratch by Heess et al.\cite{heess2017emergence} and Duan et al.\cite{duan2016benchmarking}. Nonetheless, the above works on RNN-based RL are hardly utilisable to solve locomotion POMDP in reality. This is because they do not offer the way to reuse the data collected from the far past or other agents, for training the current agent. In addition, constraints and issues on training the RNNs with small subtrajectories are barely discussed.

In this work, we examine the above two issues in a simple but reflective bipedal locomotion POMDP task on rugged terrains. Our methodology upon those challenges can successfully improve the learning performance of existing algorithms (see Table \ref{table:summary-performance} in \ref{sec:simulation}). Our contributions are below three aspects of training the RNN-based policy and value functions:
\begin{itemize}
	\item Tail-step bootstrap for interpolated Temporal Difference on Value estimator;
	\item History-aware configuration of initial hidden state;
	\item Truncation of backpropagation through time;
	\item Transfer of behavior via injecting external experience.
\end{itemize}

This paper is organized as follows. The principles of the RDPG algorithm is presented in section \ref{sec:preliminaries}. Our proposed improvement for training RDPG is elaborated in section \ref{sec:methods}. The simulation setup and simulation results are presented in section \ref{sec:simulation}, followed by final conclusion in section \ref{sec:conclusion}.

\section{Preliminaries}\label{sec:preliminaries}

\subsection{Partially observable markov decision process and long short term memory}
A POMDP is an action decision process that uses knowledge of observations and actions from previous time-steps to augment current observations, since the true state information is not available. To use DRL in walking control, a method is clearly needed for representing the previous observations; these will be useful in inferring the true current state. 

Since the true state can only be estimated via complete belief propagation from the terminal state, relaxed form of policy improvement methods via value approximation have been used, notably point-based value iteration methods \cite{Smith2005PointBasedPA}. Emergence of deep neural network models with temporal memory offered a new venue on the POMDP problem by point estimating the value that also references to the past observations. The Long Short Term Memory (LSTM) is variant of RNNs that allows the access to information further back in the history of the network via gate mechanism \cite{hochreiter1997long}. Therefore, the gated-RNNs are the widely used in DRL agents for POMDP tasks \cite{mnih2016asynchronous},\cite{heess2015memory}.


\subsection{Recurrent deterministic policy gradient algorithm}\label{sec:RDPG}

\begin{figure}[t]
	\includegraphics[width=65mm]{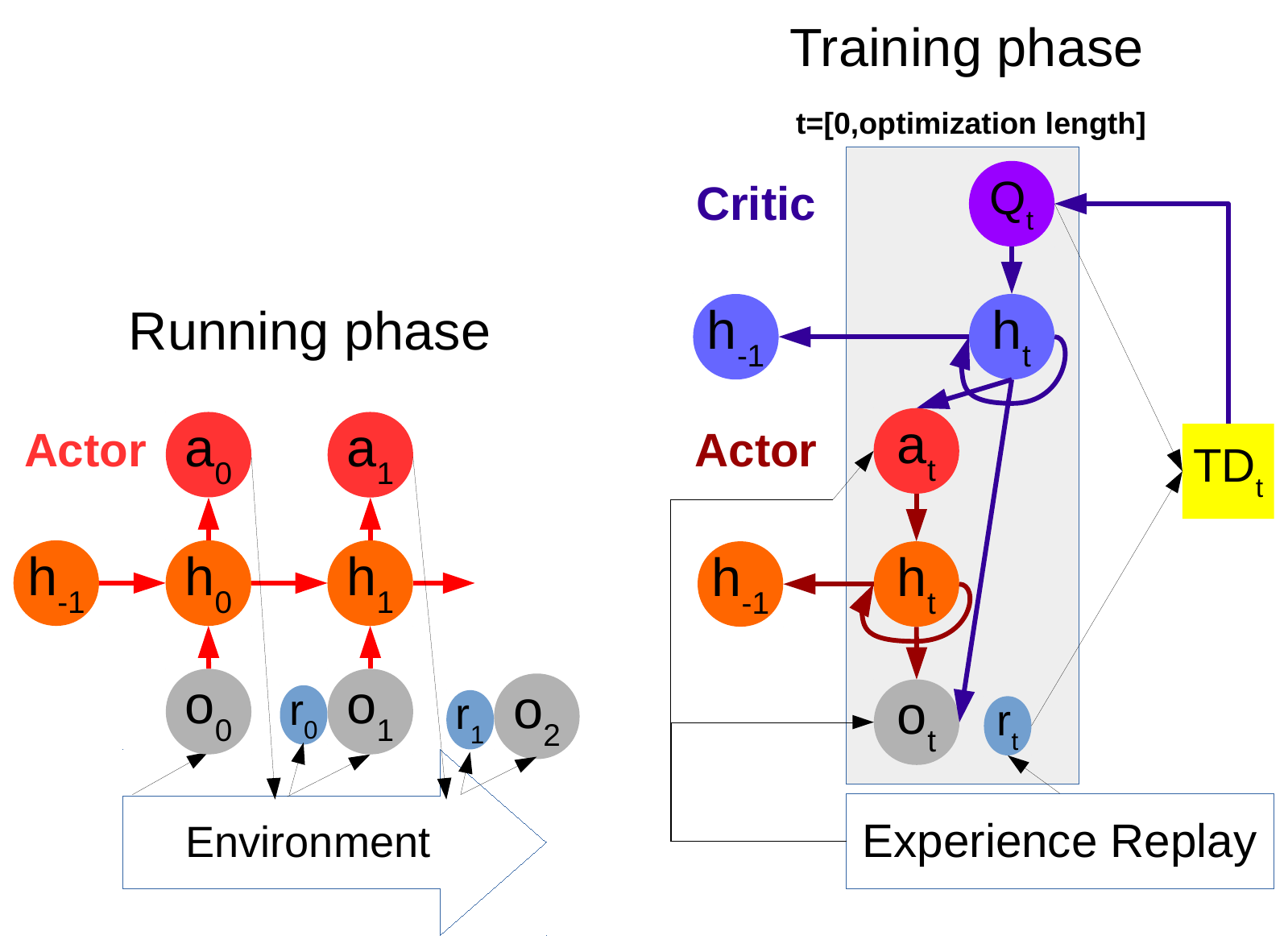}
	\caption{Recurrent deterministic policy gradient (RDPG).}
	\label{fig:rdpg_1}
\end{figure}

The RDPG is an application of Deterministic Policy Gradient algorithm \cite{silver2014deterministic} where the policy is a point-estimator of \textit{a}ction given \textit{b}elief given \textit{o}bservation built on RNNs \cite{heess2015memory}: 
\begin{align*}
\pi(a_t|(o_i,a_i)_{i=:t-1}) &= p(a_t|b_t) \approx p(a_t|o_t,h_{t-1})\\
&\equiv \delta\big(a_t-f^{Actor}(o_{0:t},h_{-1})\big)
\end{align*} 
Interpreting this as an implicit policy method of POMDP, the deterministic gradient is approximated as follows:
\begin{align}
&\bigtriangledown_{\omega}J(\pi^{\omega})\:=\:\int ds \rho(s) [\bigtriangledown_a Q^{\theta}(s,a)|_{a\sim\pi^{\omega}(s)} \cdot \bigtriangledown_{\omega}\pi^{\omega}(a|s)] \nonumber \\
&\;\;\;\approx \mathbb{E}_{s \sim \rho^{\pi^{\omega}}}[\bigtriangledown_a Q^{\theta}(o,a|h)|_{a\sim\pi^{\omega}(o,h)} \cdot \bigtriangledown_{\omega}\pi^{\omega}(a|o,h)] .
\label{eq:DPG}
\end{align}
As a member of Actor-Critic algorithm \cite{sutton2000policy}\cite{konda2000actor} with hidden state of RNN being \textit{h}istory, the Critic network approximates the state-action value and sends the score message to the Actor at policy improvement phase in the form of action gradient on the Critic's geometry ($\bigtriangledown_{a}Q^{Critic}(h,o,a)$).

In the value expectation step, the square of Temporal Difference (\textit{TD}) is minimized. \textit{TD} is an error measure between the current and target Q-values: $TD_t := Q^{target}_t - Q^{behavioral}(o_t,a_t|h_{t-1})$, where the behavioral Critic function $Q^{beh}$ estimates the Q-value. The target Critic $Q^{tar}$ is expanded to certain time-steps and then estimated by the target Critic function by Bellman equation. The original DDPG \cite{lillicrap2015continuous} and RDPG \cite{heess2015memory} uses target Q networks to provide the Q-values instead of using the learned $Q^{beh}$ networks. The weights of the target Q network is updated by slowly tracking the learned networks via ``soft updates'': $\theta^{target} = \tau\theta + (1-\tau)\theta^{target}$ by idea of non-markovian update chain.


%
The graph representation of RDPG on the POMDP framework is shown in Fig.\ref{fig:rdpg_1}. The agent's POMDP decides the action based on its \textit{belief} ($b_t$) of the true state estimated by the \textit{history} ($h_{t-1}$) representation \cite{heess2015memory} of the past observation-action trajectory and the current \textit{observation} of the state ($o_t$).


In order to facilitate exploration of new behaviousr, DPG algorithm noises its action via first order stochastic process such as Ornstein-Uhlenbeck process. In addition, we peroidically noise to the parameter space of the Actor\cite{plappert2018parameter}.


\subsection{Minibatch sampling from experience replay buffer}\label{sec:experiencereplay}
\begin{figure}[t]
	\captionsetup{justification=justified}
	\includegraphics[width=70mm]{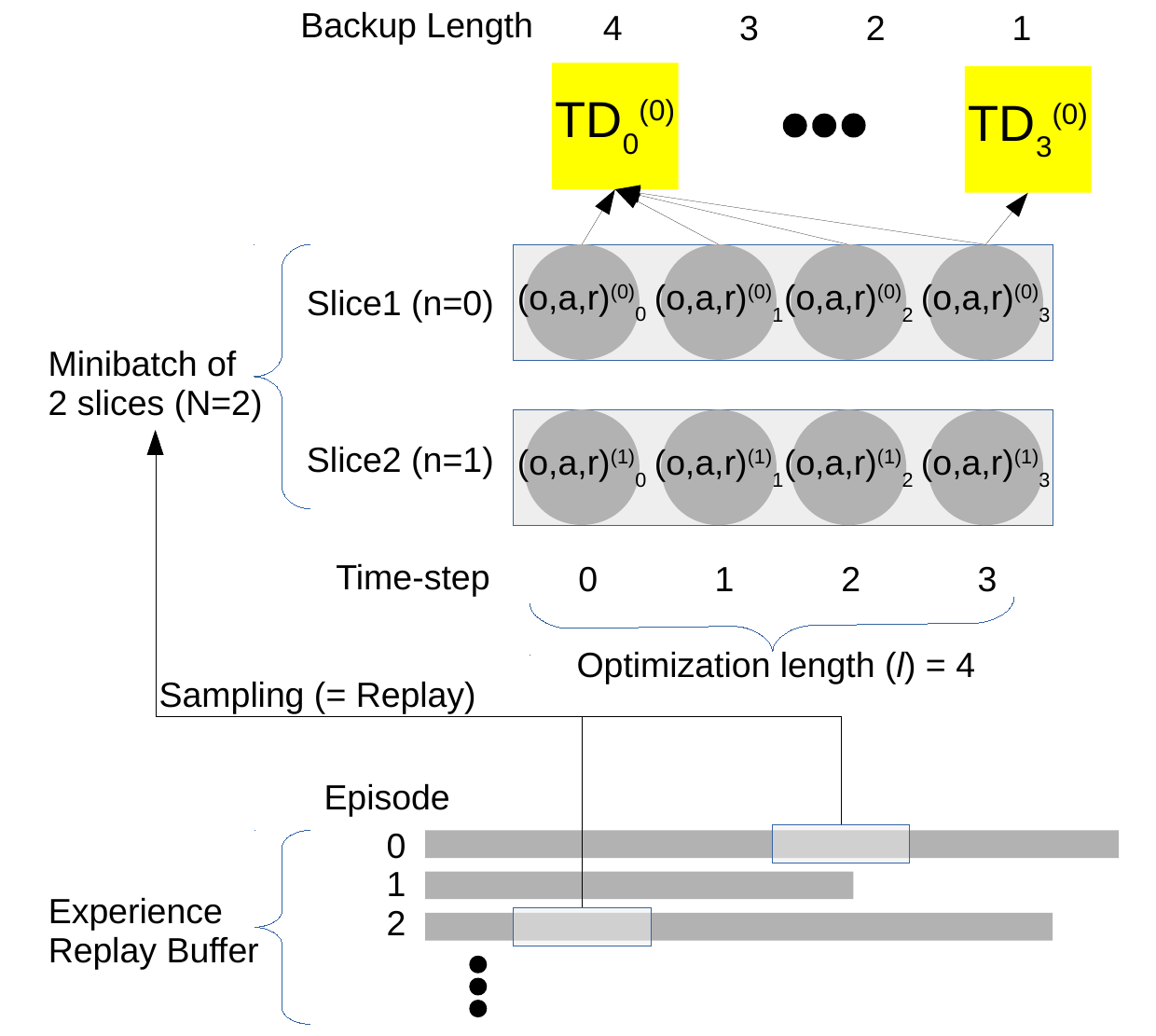}
	\caption{Episodic \textit{Experience replay} buffer for training.}\label{fig:experiencereplay}
\end{figure}

All the pairs of state transition, action, and reward experience are saved in an \textit{Experience replay} buffer \cite{lin1992self} for the minibatch gradient update during the running phase. The \textit{Experience replay} buffer also provides training data in the form of minibatch. Since this technique decorrelates the training data from the data collected on the current episode, it has been preferred over on-line \cite{mnih2016asynchronous} or episodic-wise learning methods \cite{hausknecht2015deep} by recent works on DRL \cite{mnih2015human}\cite{heess2015memory}\cite{lillicrap2015continuous}.

In our model, the buffer stores the experience pair at each time-step $(o,o',a,r)_t$ and forms a bundle of episodic trajectories (Fig.\ref{fig:experiencereplay}) so that the RDPG model will not be trained with slices having terminal state in the middle. The episodic buffer provides the minibatch of experience for training by following steps. First, the agent picks episodes to sample. Second, the agent samples one slice of certain length $l$ each from the chosen episode, and stacks the slices as a minibatch. Lastly, the RNN-based Actor and Critic networks read the entire slices, measure their objectives, and generate the update gradient of parameters for the Policy Iteration.

A major challenge in applying the minibatch-based training method to the RNN-based RL agents is that the training data is fed with only a fraction of each episode while the RNN memorizes each of the entire episode during the run.

\begin{algorithm}[t]
	\SetKwInOut{Input}{Input}
	\SetKwInOut{Output}{Output}
	
	\textbf{def RDPG}:\\
	
	Initialise $\omega_{\texttt{Actor}}, \theta_{\texttt{Critic}}, B_{\texttt{experience replay}}$\\
	Set optimization/update/scan length as $l/u/s$ $\;\;\#\; l\geq u$ \\
	\For{episode=1 to M}{
		Initialise OU-noise $\mathcal{N}_{epi}$ and Noise() parameters\\
		Initialise observation $o_0$, and hidden states $h_0$ \\		
		\While{not terminated}{
			Generate action: $a_t = \pi^{\omega}(o_t,h^{Actor}_{t-1}) + \mathcal{N}_{epi,t}$\\
			Receive $o_{t+1}, r_t$ from the environment($o_t, a_t$)\\
			Store transition($o_t,a_t,r_t,o_{t+1}$) to $\textit{B}$\\                
		}        
		Denoise() parameters and then Update()\\		
	}    
	\BlankLine    
	\textbf{def Update}:\\
	Sample $\textit{N}$ sliced trajectories of length $s+l$ from $\textit{B}$\\
	Get $h_{-1}$ from $h_{init}$ by scanning $s$ time-steps (Eqn.\ref{eqn:scan})\\
	Get TD gradient of the minibatch (Eqn.\ref{eq:unsyncedinterpolatedTD}): 
	$\bigtriangledown_{\theta_{old}}L = -\frac{1}{\textit{N}} \overset{N}{\underset{n=1}{\sum}}\Big((\lambda \frac{1-\lambda^{u}}{1-\lambda})^{-1} \overset{u-1}{\underset{i=0}{\sum}} \lambda^{i}\cdot{}^{(l-i)}TD^{(n)}_{i} \cdot
	\bigtriangledown_{\theta_{old}}Q^{\theta}(o^{(n)}_{i},a^{(n)}_{i}|o^{(n)}_{0:i},h_{-1}^{(n),Critic}) \Big)$\\ 	    
	Update Critic by minimizing the TD loss with ADAM optimizer\cite{kingma2014adam}: $\theta_{new} \leftarrow ADAM(\theta_{old},\bigtriangledown_{\theta_{old}}L)$\\
	Get Actor gradient of the mini-batch (Eqn.\ref{eq:DPG}): $\bigtriangledown_{\omega_{old}}J = \frac{1}{\textit{N}}\overset{N}{\underset{n}{\sum}}\Big(l^{-1}\overset{l-1}{\underset{i=0}{\sum}}\bigtriangledown_a Q^{\theta}(o^{(n)}_{i},a|h^{(n),Critic}_{-1+i})|_{a\sim\pi^{\omega}(o^{(n)}_{i},h^{(n),Actor}_{-1+i})} \cdot \bigtriangledown_{\omega_{old}}\pi^{\omega}(a|o^{(n)}_{i},h^{(n),Actor}_{-1+i})\Big)$\\
	Update Actor: $\omega_{new} \leftarrow ADAM(\omega_{old},\bigtriangledown_{\omega_{old}}J)$\\
	\caption{Recurrent Deterministic Policy Gradient}\label{alg:RDPG}
\end{algorithm}

\section{Learning methods}\label{sec:methods}
Following the discussion in Sec. \ref{sec:RDPG} and \ref{sec:experiencereplay}, our work focus on issues of Policy Iteration process of RDPG while the update gradients are calculated from sampled minibatches. A pseudocode of our model RDPG algorithm is detailed in Algorithm \ref{alg:RDPG}.

\subsection{Modifications on the optimization of RDPG}
\begin{figure}[t]
   \centering   
   \begin{subfigure}[t]{0.44\textwidth}
       \includegraphics[width=\textwidth]{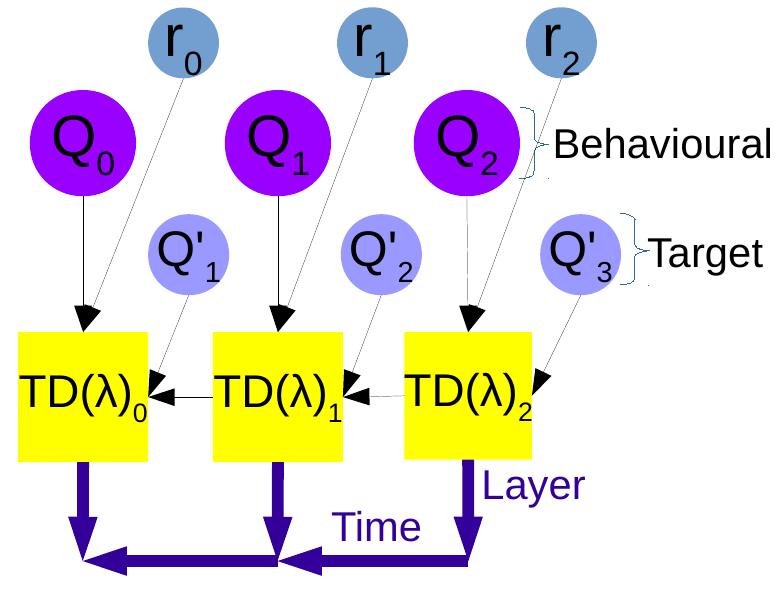}
       \caption{$TD(\lambda)$ backup\cite{harb2017investigating}}
       \label{fig:nstep_2}
   \end{subfigure}
   ~ 
   \begin{subfigure}[t]{0.36\textwidth}
       \includegraphics[width=\textwidth]{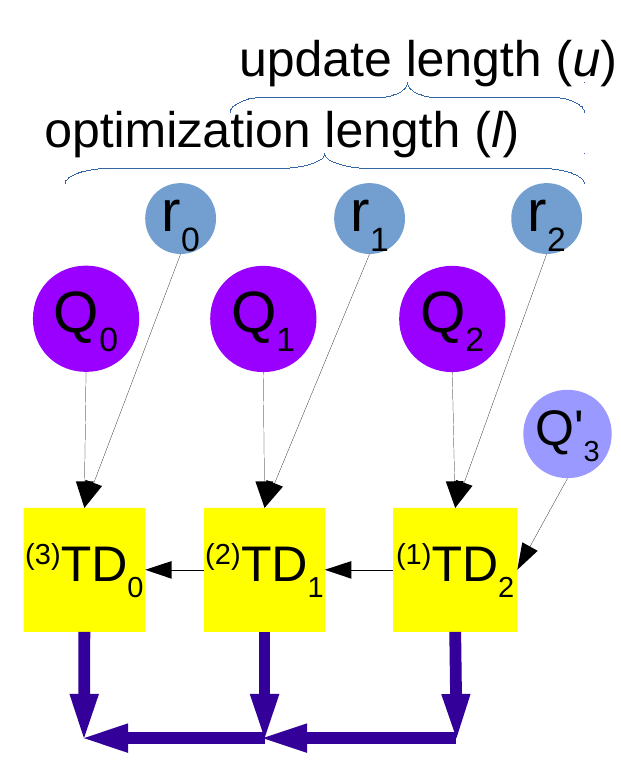}
       \caption{Tail-step $TD(\lambda)$}
       \label{fig:nstep_3}
   \end{subfigure} 
   \caption{TD measurement in a subtrajectory. Thick purple line: backpropagation of Critic update gradient.}\label{fig:nstep_design}  
\end{figure}

\subsubsection{Tail-step bootstrap of interpolated $TD$}\label{sec:nstep}
The original RDPG algorithm proposed by Heess et al. \cite{heess2015memory} measured one-step \textit{TD} for every experience pair within the minibatch. The problem with the one-step TD is that the obtained reward can only be used to directly update the value of the specific state action pair since the values of other state action pairs can only be updated indirectly through the Q-value function Q(s, a). If the observation of environment (o) does not truly represent the true state (s) such as POMDP tasks, the short backup length induces the bias of the estimated Q-value. This justifies the necessity of the Critic referring to farther time-steps.

Giving the longest but equal backup length for every \textit{TD} measures of sampled experience pairs is crucial for theoretically justifying the Q-value's convergence bound. However, minibatch sampling shortens down the backup length for later experience pairs within the sampled subtrajectories (See Fig.\ref{fig:experiencereplay}). Unless the full Monte-Carlo sums of future rewards are utilised to update Critic, which is not the case for the DPG algorithm, the problem sustains. A common practice is to ignore the equality on length but to backup with truncated multi-step TD's:
$${}^{(i)}TD_{t}=\overset{t+i-1}{\underset{t'=t}{\sum}} \gamma^{t'-t}r_{t'}+\gamma^{i} Q^{tar}_{t+i} - Q^{beh}(o_t,a_t|h_{t-1})$$ 
is the multi-step \textit{TD} \cite{Sutton1998} with backup length $i$. The length $i$ varies by the position of experience pairs within the sampled subtrajectories. This raises another issue on the variance control of sampled rewards sum for each experience pair due to uneven backup length, resulting in lack of principled appreciation on recency of TD's\cite{schulman2015high}.

In terms of bias control, recent advances either refer to the whole trajectories \cite{schulman2015high} or boostrap impractically long temporal horizon \cite{schulman2017proximal}, or train the agent episodically online from backward direction\cite{mnih2016asynchronous}. They are not suitable for subtrajectory sampled training. In addition, the control on variance was barely discussed.

Our method is motivated from the work of Harb et al. for use of $TD(\lambda)$ in the context of RNN-based Q-learning \cite{harb2017investigating} (Fig.\ref{fig:nstep_2}). Although the $TD(\lambda)$ measurement improved the learning performance, their work 1) did not resolve the problem of shortening backup length of the $TD$'s in the later time-steps of subtrajectories, and 2) discrete action control tasks can explore much diverse action values for similar states since action representation is categorical. 

In order to ensure the maximum backup length for every experience pair in the minibatch, we propose a new method of updating the Critic. First, we generalize the \textit{TD} measurement as an interpolation of multi-step \textit{TD}'s:
\begin{align}
Z^{-1}\overset{l}{\underset{i=1}{\sum}}w(i) \cdot {}^{(i)}TD_t \text{, where } Z^{-1}\overset{l}{\underset{i=1}{\sum}}w(i)=1 \label{eq:interpolatedTD}
\end{align}
On the other hand, our method enables a principled backup scheme while still measuring only one \textit{TD} with the longest backup length each experience pair (Fig.\ref{fig:nstep_3}). 

In our method, the target Q-value is only being calculated at the end of each slice of sampled trajectories, which in turn measures $TD(\lambda)_{\lambda=1}$ for every point. Hence each slice with length \textit{l} measures
$\{{}^{(l-x)}TD_{t+x}\}_{x=0\cdots l-1}$ as that of $TD(\lambda)_{\lambda=1}$. However, an interpolated-$TD_t$ with backup length \textit{l} requires $\{{}^{(i)}TD_{t}\}_{i=1\cdots l}$. 
They correspond to each other as follows: 
\begin{align*}
\{{}^{(l-x)}TD_{t+x} \text{ is measured for }interpolated\text{-}TD_{t+x}\}_{x \in [0,l-1]} 
\end{align*}
Without waiting for the other elements of the \textit{interpolated-TD}'s \footnote{For instance, we still need $\{{}^{(i)}TD_{t+i}\}_{i=1\cdots,i-1,i+1, \cdots l}$ to estimate the $interpolated\texttt{-}TD_{t+i}$ since only one ${}^{(l-i)}TD_{t+i}$ out of constituents is measured in the sampled subtrajectory.}, the agent updates the Critic with the part of them we measured on the minibatch, which are $\{{}^{(l-x)}TD_{t+x}\}_{x=0\cdots l-1}$.

However, all the other constituents will eventually be visited by sufficient rounds of replay with slices of experience from different location\footnote{We ensure the eveness of visitation by uniformly sampling subtrajectories and episodically saving and removing trajectories.}. Therefore, every experience points gather all the multi-step \textit{TD}'s as constituents for the $l$-length interpolated-$TD$ in the end.

To minimize the bias of our interpolation in fixed $l$, we apply decaying weight factor $w(i)=(\lambda \frac{1-\lambda^{l}}{1-\lambda})^{-1}\lambda^{i}$ (Eqn.\ref{eq:interpolatedTD}) to ${}^{(i)}TD$'s , resulting in the interpolated-$TD_t$ as $(\lambda \frac{1-\lambda^{l}}{1-\lambda})^{-1}\overset{l}{\underset{i=1}{\sum}}\lambda^{i}\cdot{}^{(i)}TD_t$. This is an adoptation of $TD(\lambda)$ \cite{sutton1988learning} in a fixed sample length while both action and state visitation conditions are ignored, since our agent learns from contiuous state-action space. Eventually, the tail-step bootstrap of the interpolated-$TD$ to the Critic becomes:
\begin{align}
(\lambda \frac{1-\lambda^{l}}{1-\lambda})^{-1}\overset{l-1}{\underset{i=0}{\sum}}\lambda^{i}\cdot {}^{(l-i)}TD_{t+i}
\label{eq:unsyncedinterpolatedTD}
\end{align}
per sliced trajectory (subtrajectory) sampled on a minibatch. When $\lambda = 1$, our method becomes equivalent to the truncated multi-step $TD$ bootstrapping where $w(i)=1\;\;\forall i$, which shows that this bootstrapping method actually trains the Critic in interpolated manner upto fixed optimization length.

Throughout experiments, we search for the optimum backup length (update length) for the bootstrapping conserved for every experience pairs. Lastly, we also investigate on the optimal length to backpropagate the update gradient of LSTM-Critic through time (optimization length: see Fig.\ref{fig:nstep_3}).

\subsubsection{Hidden state initialization via trajectory scanning}\label{sec:scan}
\begin{figure}[t]
	\captionsetup{justification=justified}
	\includegraphics[width=0.95\textwidth]{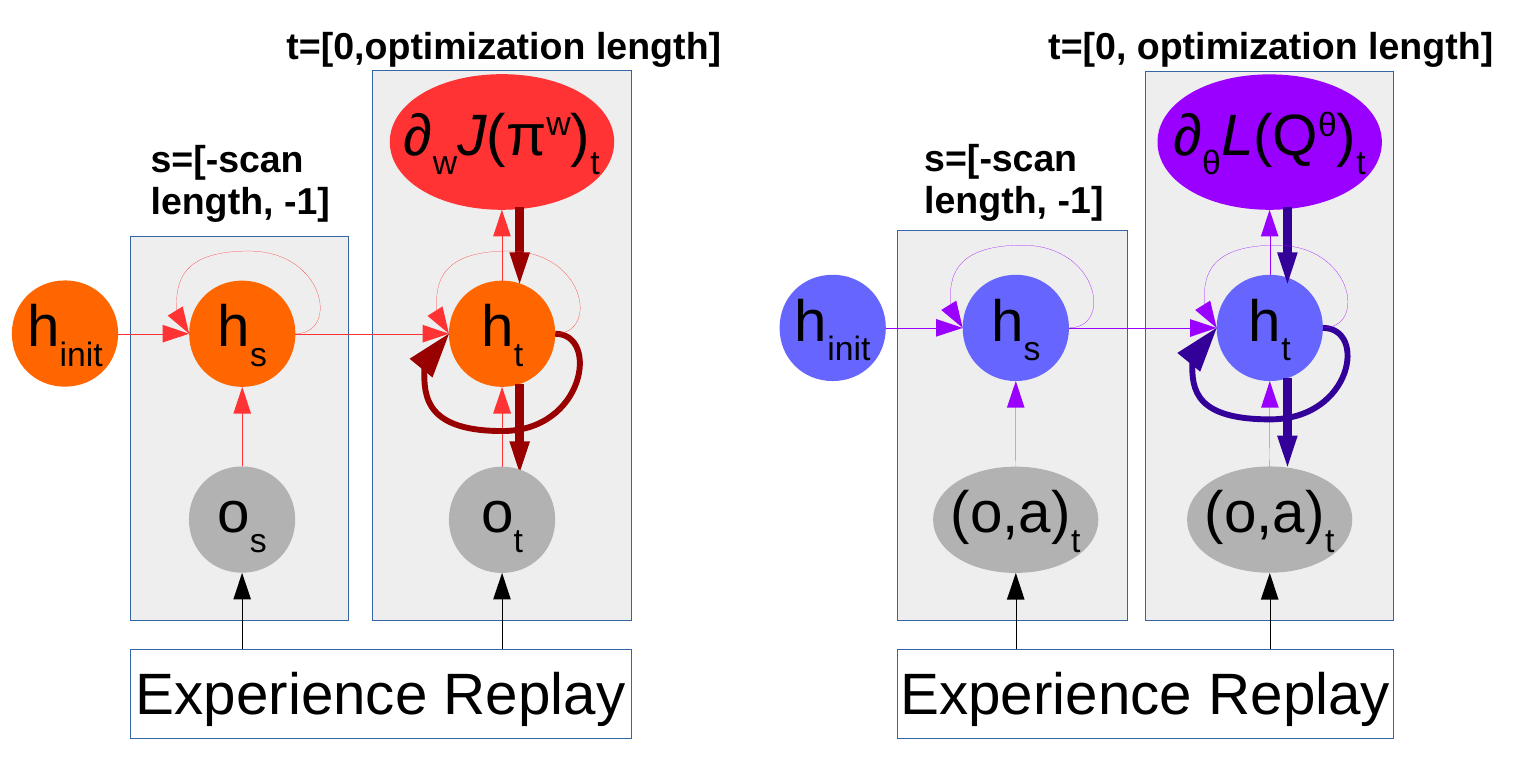}\caption{Scanning strategy that looks back from constant $h_{init}$ to $h_{-1}$ on Actor (left) and Critic (right) during training. Without the scanning interval, $h_{-1}$ is equal to $h_{init}$.}\label{fig:init_h}
\end{figure}

One limitation of the minibatch gradient update for RNN-based DRL is: the historic representations of the networks during the training phase no longer accurately reflect the truth history obtained from the real trial. This is because the sliced trajectories of minibatches decorrelate from the past state-action pairs within the episodes, hence the initial hidden states $h_{-1}$ of the slices become completely \textit{unaware} of the episodic history. On the other hand, the episodic-wise learning method is utilized since each training data is read from the first or last time-step of its episode \cite{mnih2016asynchronous}\cite{hausknecht2015deep}, and the training data is highly correlated to each other.

However, the importance of having meaningful initial hidden states has been actively searched since the LSTM is capable of memorizing useful long-term history. We aim to develop a method to enforce the $h_{-1}$ to represent meaningful history while the minibatch cuts off the past trajectories of the slices (Fig.\ref{fig:experiencereplay}).

We introduce a recent technique mitigating this issue in discrete control domain: scanning the past trajectory before the optimization interval (Fig.\ref{fig:init_h})\cite{harb2017investigating}\cite{lample2017playing}. By scanning, $h_{-1}$ represents the history ahead of the interval by: 
\begin{align}
h^{Actor}_{-1}&=f^{Actor}\big((o_t)_{t=-scanning\;length:-1}|h_{init}\big)\nonumber\\
h^{Critic}_{-1}&=f^{Critic}\big((o_t,a_t)_{t=-scanning\;length:-1}|h_{init}\big)\label{eqn:scan}
\end{align}
where $f$ is the recurrent part of Actor or Critic representing the past trajectory as a history vector, and $h_{init}$ is a zero vector. The scanning interval’s solely purpose is to generate $h_{-1}$ and does not generate any update gradient. In this paper, we explicitly validate the scanning strategy in terms of its compatibility on the Actor-Critic framework for continuous action decision task by measuring learning performance as well as analysing locomotion behaviour.


\begin{figure}[t]
	\captionsetup{justification=justified}
	\includegraphics[width=55mm]{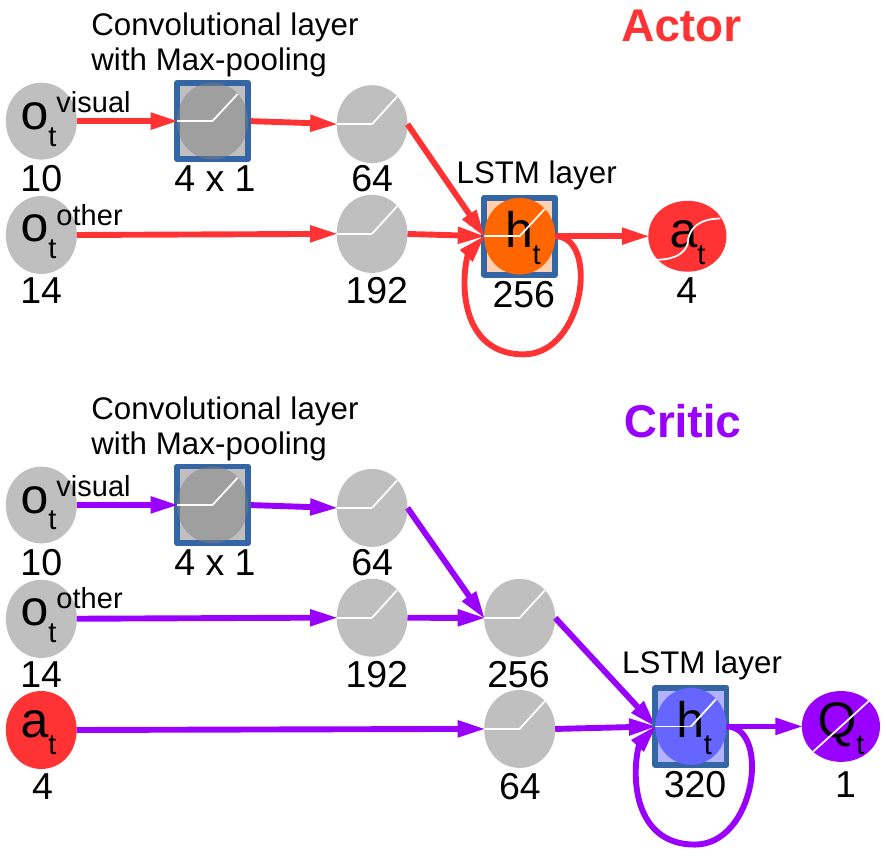}
	\caption{Network design of Actor and Critic.}\label{fig:network}
\end{figure}

\subsubsection{Truncation of temporal flow of update gradient}
During the policy improvement step, we completely truncate out the temporal Backpropagation path (BPTT) of the LSTMs. Without BPTT, the estimated action at each time-step only refers to the current Q-values for applying the policy gradient $\bigtriangleup \omega_t = \bigtriangledown_{a_t} Q_t(h_t,o_t,a_t)$ $\cdot \bigtriangledown_{\omega} \pi(h,o_t)|_{h=h_t}$. If BPTT is enabled, then the update gradient of Actor becomes $\bigtriangleup \omega_t = \sum_{t'' \leq t' \leq t} \bigtriangledown_{a_{t'}} Q_t(h_t,o_t,a_t)$ $\cdot\bigtriangledown_{\omega} \pi(h_{t''},o_{t''})$. The resultant policy gradient given $o_t$ refers to Q-values in future which already encompasses the present value of all the expected rewards from present to future. This issue is only addressed in context of online supervised learning \cite{ollivier2015training}.

\subsubsection{Experience injection for behaviour transfer}\label{sec:injection}
From prior simulations, we have found that the RDPG agents tend to learn monotonic behaviors. \textit{Monotonic} behavior means that the agent sticks to one pattern of action behavior (or motion) to solve every category of landscape. Monotonic behavior is bad for traversing complex terrains as it is necessary to have different behaviors to negotiate different terrain features. 

We propose a method called \textit{experience injection} to transfer knowledge between agents to resolve the issue of monotonic behavior. In our proposed experience injection, knowledge is transferred from source agents (teachers) to a recipient agent (student) in the form of trajectories of state-action-reward pairs. Concretely, the experience injection provides the interpolated-$TD_t$ measure (Eqn.\ref{eq:interpolatedTD}) to the student's Critic as:
\begin{align*}
&\overset{l}{\underset{i=1}{\sum}}w(i) \cdot {}^{(i)}TD^{\theta_{student}}_t \big|_{a \sim \underset{j \in teachers,student}{\cup} \pi^{j}(o),\; (o',r) \sim env(o,a)}
\end{align*}

By injecting trajectories of well-trained agents into the experience replay buffer of a new agent, we can introduce good learned behavior into the new agent so that it can learn more optimal behavior by deciding non-monotonic action chain. Teachers can be trained from various policy learning algorithms, network designs, and even reward designs before injection. Normalization issue of multiple teacher policies can be ignored since DPG is an off-policy learning algorithm. However, we do not dominate the experience replay buffer with external trajectories since the trajectory distribution distorts \cite{zhang2017deeper} and anneal down the contribution of injected trajectories over episodes.

The idea of injecting external experience into an agent is well examined in context of inverse reinforcement learning \cite{finn2016guided}. However, our injection method does not limit experiences to be bounded near to the current policy of student. Policy distillation \cite{rusu2015policy} property of experience injection method is justified since the Critic is shaped with diverse behaviours of multiple teachers, and then the Actor maximises Q-value with guidance of its Critic.

\subsection{Network structure}

\begin{figure}[t]
	\captionsetup{justification=justified}
	\includegraphics[width=55mm]{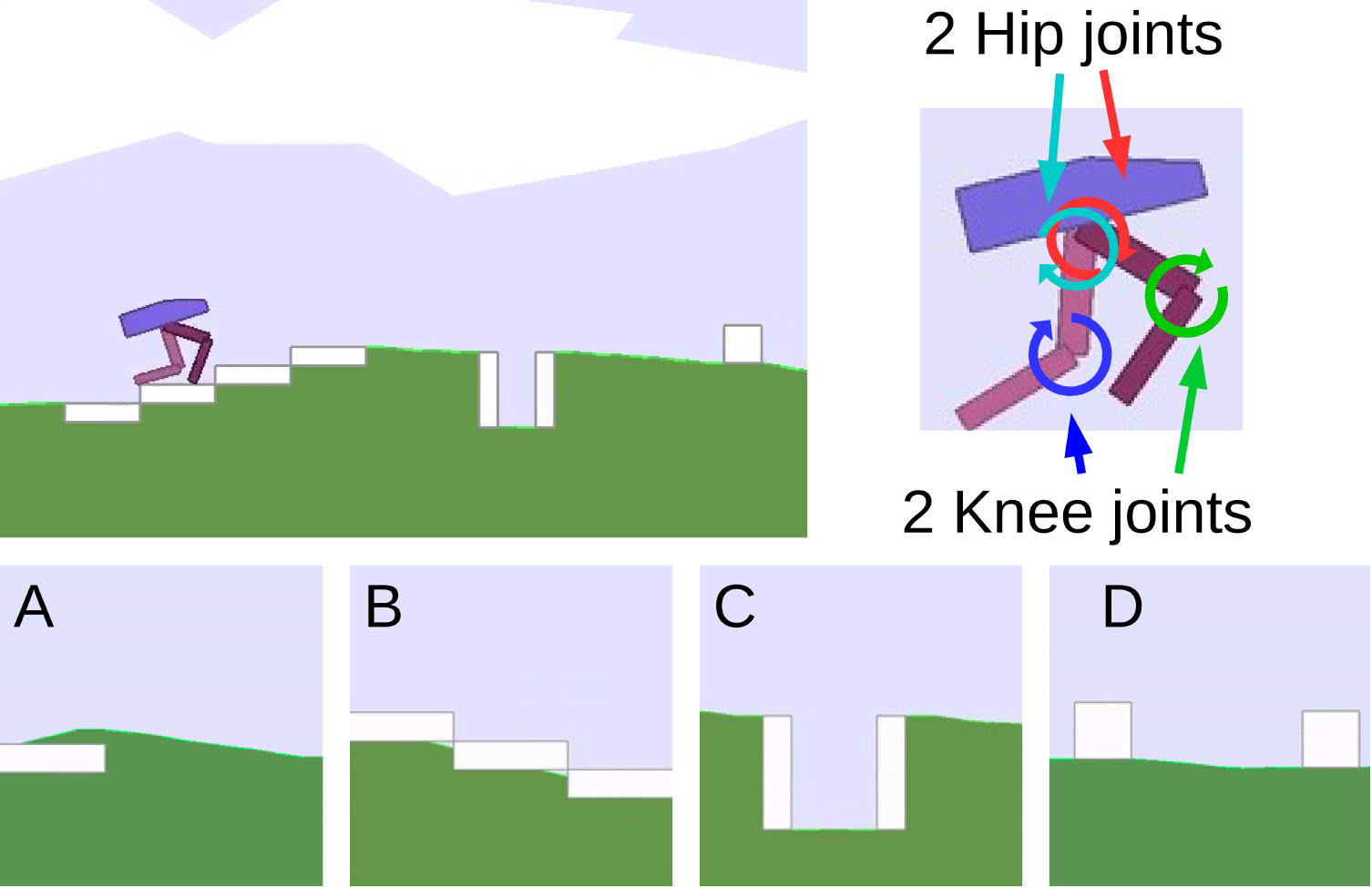}\caption{Terrain feature. A. Slope. B. Stair. C. Gap. D. Hurdles.}\label{fig:hard}
\end{figure}

We design\footnote{Throughout experiments with various network designs, the performance nor behaviour of RDPG does not vary significantly.} the Actor (Fig.\ref{fig:network}) as a four-layer network. The first layer transforms the visual observation with 64 channels of convolutional filters followed by max-pooling. The second layer converts the first layer and the other non-visual observation with 64 and 192 feedforward neurons, respectively and then concatenates them. The second layer is the LSTM of 256 neurons with Relu. The top layer takes feedforward neurons with tanh for 4-dimensional(D) action output. The output of the Actor network is a 4-D torque reference for knee-hip joints of both legs (Fig.\ref{fig:hard}). The tanh activation function is to limit the output range within [-1,1].

The Critic (Fig.\ref{fig:network}) shares the structure of the Actor but with one more feedforward layer between the second and the third layer. The additional layer augments the 4-D action output to 64 neurons and then concatenates with the second layer. Moreover, we increase the size of LSTM layer of the critic network to 320 neurons. The top layer of the Critic uses linear identity activation instead of tanh activation.

%

\section{Simulation study}\label{sec:simulation}
\subsection{Simulation environment setup}
\subsubsection{OpenAI Gym BipedalWalker environment}

The bipedal walking task that we aim to solve is the OpenAI's Bipedal-Walker challenge\footnote{https://gym.openai.com/envs/BipedalWalkerHardcore-v2}. The 2D simulation environment is partially observable to the bipedal walking agent.

The bipedal character has 4 degrees of freedom, 2 hip and knee joints. The simulation environment provides 24-dimensional sensory feedback. This information consists of 10-D LIDAR (visual) with limited range, 4-D translational/rotational displacement and velocity of hull, 8-D rotational displacement and velocity of the joints, and 2-D binary contact of the feet terrain. The control loop runs the same frequency as the physics simulation at 50Hz.

The goal of the challenge for the bipedal agent is to traverse a variety of rugged terrains (Fig.\ref{fig:hard}) for 360 points without falling. The environment runs episodically, ie an episode terminates if: the body of robot touches the environment, or the agent reaches the goal, or the maximum runtime (40s) is out.

\subsubsection{Reward design}
%

Our reward design solely focuses on facilitating the agent to move as quick as possible ($\vartriangle x_t/\vartriangle t$) with mild penalty against collision to the ground ($r_{c,t}=-20$): $$ r_t = \frac{\vartriangle x_t}{\vartriangle t} + \delta(\text{collision}_t==\text{True})\cdot r_{c,t}$$ 

Although the resultant behaviour is less realistic due to high torque of actuators, we found that posture and stability penalties significantly deter the emergence of gait behaviours due to the partial observability, frequency and difficulty of the obstacles (See Table \ref{table:summary-performance}).

\begin{figure}[t]
	\centering   
	\begin{subfigure}[t]{\textwidth}
		\includegraphics[width=\textwidth,trim= 0cm 0.6cm
		0cm 0cm, height=2.7cm, clip]{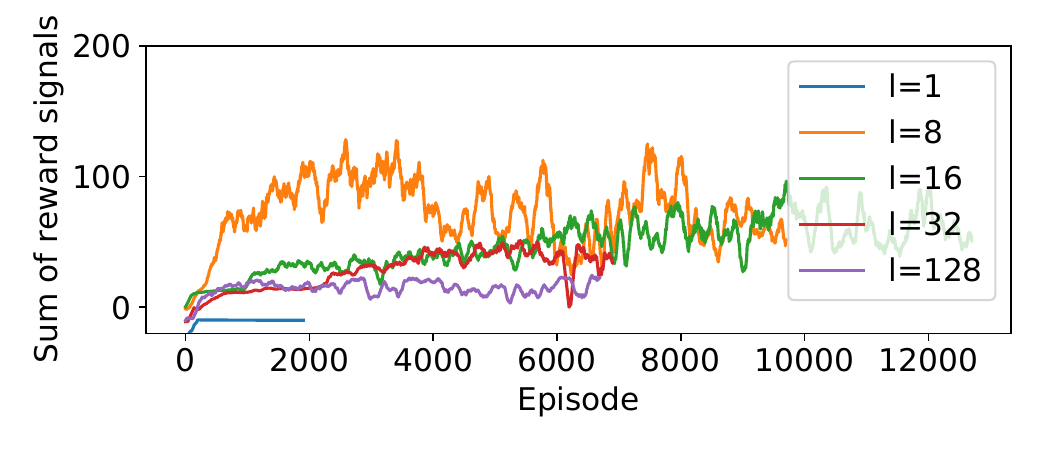}        
		\caption{Learning curves of agents with various optimization length ($l$). Update length ($u$) is set as half of the optimization length.}
		\label{fig:optlen_hard}
	\end{subfigure}
	
	\begin{subfigure}[t]{\textwidth}
		\includegraphics[width=\textwidth,trim= 0cm 0.6cm
		0cm 0cm, height=2.7cm, clip]{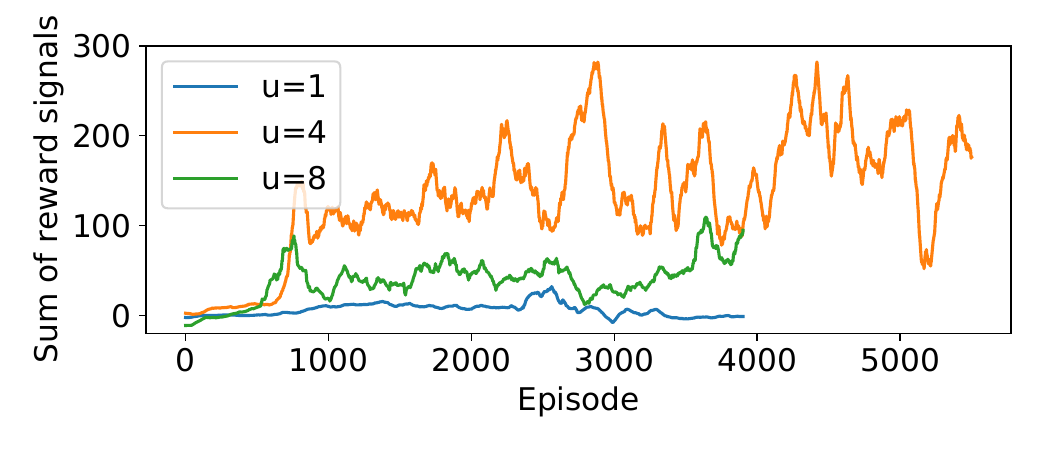}
		\caption{Learning curves of agents with various update length and fixed $l$ as 8. The agents are trained on the terrain without obstacles otherwise the agents with $u$=1 or 8 do not learn to walk.}
		\label{fig:optupdlen_easy}
	\end{subfigure}
	\caption{Learning performance of agents with various optimization and update length for the interpolated TD method.}
	\label{fig:multi-nstepTD}
\end{figure}

\begin{figure}[t]
	\centering   
	\begin{subfigure}[t]{\textwidth}
		\includegraphics[width=\textwidth,trim= 0cm 0.6cm
		0cm 0cm, height=2.7cm, clip]{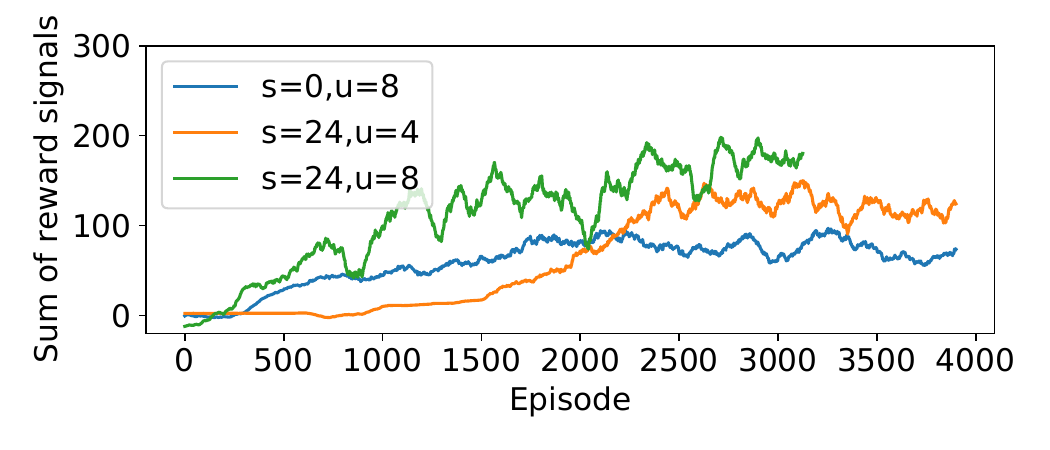}        
		\caption{Learning curves of agents with various scanning interval lengths ($s$). Optimization length is set as 8.}
		\label{fig:scan}
	\end{subfigure}
	
	\begin{subfigure}[t]{\textwidth}
		\includegraphics[width=\textwidth,trim= 0cm 0.6cm
		0cm 0cm, height=2.7cm, clip]{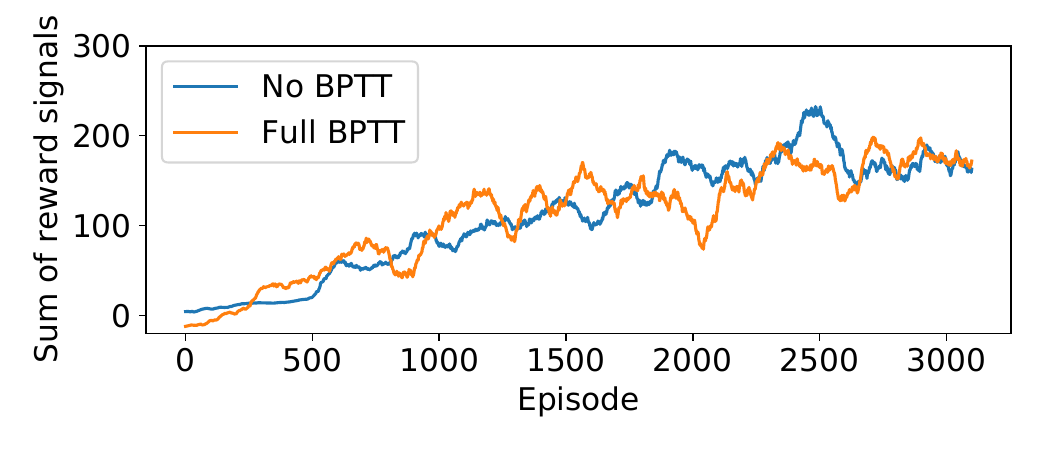}
		\caption{Effect of BPTT truncation on learning performance. Same hyperparameters are set as $u=l=8,  s=24$.}
		\label{fig:bpttTrun}
	\end{subfigure}
	\caption{Scanning initialisation and BPTT control.}
	\label{fig:scanAndBPTTtrun}
\end{figure}

%

\subsection{Simulation results}

\begin{figure*}[t]
	\centering		
	
	\begin{subfigure}[t]{0.8\textwidth}
		\includegraphics[width=\textwidth]{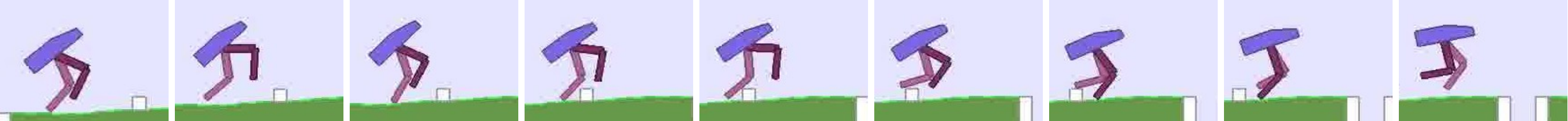}
		\caption{Walking on flat terrain with a bump}
		\label{fig:flat}
	\end{subfigure}	
	\begin{subfigure}[t]{0.8\textwidth}
		\includegraphics[width=\textwidth]{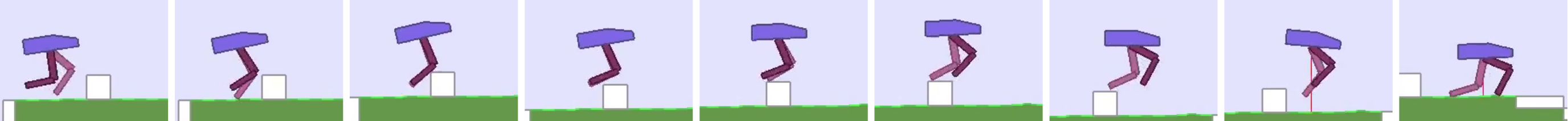}
		\caption{Jumping over a hurdle with stable landing}
		\label{fig:hurdle}
	\end{subfigure}
	
	\begin{subfigure}[t]{0.8\textwidth}
		\includegraphics[width=\textwidth]{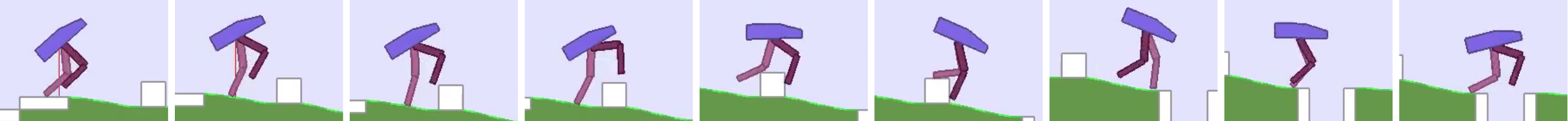}
		\caption{Crossing over a hurdle followed by agile adjustment of posture to jump over a gap}
		\label{fig:hurdle_gap}
	\end{subfigure}
	
	\begin{subfigure}[t]{0.8\textwidth}
		\includegraphics[width=\textwidth]{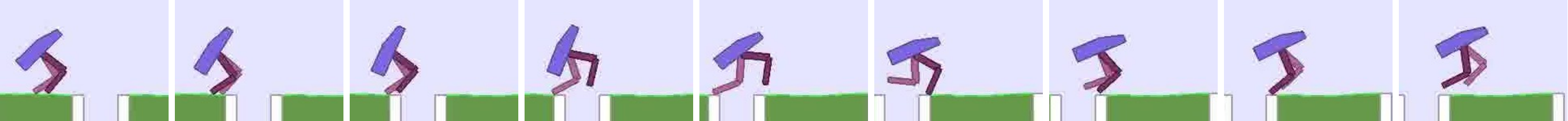}
		\caption{Leaping over a gap}
		\label{fig:gap}
	\end{subfigure}
	
	\begin{subfigure}[t]{0.8\textwidth}
		\includegraphics[width=\textwidth]{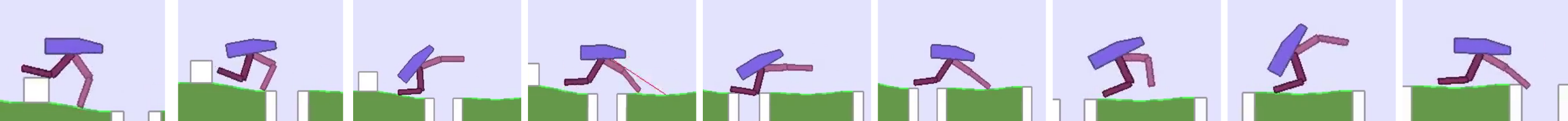}
		\caption{Crossing over a gap: a pure RDPG behaviour without experience injection}
		\label{fig:gap2}
	\end{subfigure}
	
	
	\begin{subfigure}[t]{0.8\textwidth}
		\includegraphics[width=\textwidth]{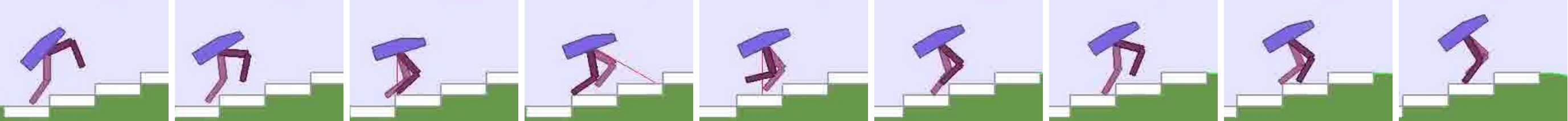}
		\caption{Walking over stairs}
		\label{fig:stair}
	\end{subfigure}	
	
	\caption{Terrain specific agile behaviors generated by our RDPG agent trained with experience injection.}\label{fig:behavior}
\end{figure*}

Our DRL framework has successfully trained policies for a stable and dynamic locomotion\footnote{Video is available at https://youtu.be/ijU4MfdaF8k\label{fn:video}}. It is capable of negotiating a diversity of terrain features including slopes, stairs, gaps and hurdles in a very agile manner. We have also benchmarked our modified RDPG with the existing Feedforward network-based Policy gradient algorithms including DPG and KL-divergence controlled Proximal Policy Optimization algorithm (PPO$_{KL}$)\cite{schulman2017proximal}.

Learning performance is measured with R(100MA) - episodic sum of rewards averaged over the last 100 episodes - and Success ratio - percentage of episodes out of 200 the best agent reaches the goal without fall. The best learning performance of the models are summarized in Table \ref{table:summary-performance}. 
Snapshots of locomotion of our RDPG agent are featured in Fig.\ref{fig:behavior}.

\begin{table}[ht]
	\centering
	\caption{Summary of performance of neural DPG.}
	\begin{tabular}{ m{4.3cm}  m{1.3cm} m{1.0cm} } 
		\hline
		Model & Top \texttt{$R_{100MA}$} & Success({$\%$})\\ 
		\hline\hline
		Our RDPG \footnote{\label{fn:ours}Optimization/Update/Scanning length set as 8/8/120, and no BPTT} & 227.5 & 23.5 \\ 
		\hline
		Our RDPG with injection & \textbf{238.5} & \textbf{32.5} \\
		\hline
		Baseline RDPG\cite{heess2015memory}\footnote{TD(0) backup without scanning: N/A means the agent does not walk} & N/A & 0.0 \\
		\hline
		DDPG\cite{lillicrap2015continuous}\footnote{Network spec is same as Fig.\ref{fig:network} except the LSTM layer being feedforward} & 194.6 & 28.0\\ 
		\hline
		DDPG with injection & 187.0 & 11.5\\		
		\hline
		FFN-$PPO_{KL}$\cite{heess2017emergence} & 72.7 & 0.5 \\
		\hline
	\end{tabular}\label{table:summary-performance}
\end{table}


\subsubsection{Optimal TD backup interval search in interpolation method}
We grid-search the optimal optimization length and the ratio of update length. Fig.\ref{fig:multi-nstepTD} shows that the optimality of policy pursues balance on the variance of $Q^{\pi}_{Critic}$ over the range of interval lengths.

As mentioned in Section \ref{sec:nstep}, longer optimization length results in the Critic network estimating discounted rewards in farther time-steps rather than nearer ones; hence the Critic produces a state-action value that is less correlated to the current time-step. Shortening the update length to one time-step hampers learning performance, as can be seen in the figure. When the update length is 1 so that the interpolated-TD is equal to the 1-step TD backup, the agent is not able to learn a successful walking policy as in Fig \ref{fig:optupdlen_easy} that the performance converges within a few hundreds of episodes and no longer improves. 

\subsubsection{Initializing hidden state via trajectory scanning improves motion behavior}
Fig.\ref{fig:scan} shows the scanning strategy facilitates better learning performance as well as training speed of the agent. Scanning the past trajectory may provide a more reasonable initial hidden LSTM state to both the Actor and Critic network. This results in improved learning performance due to the reduced bias of the Q-value estimated by the RNN-based Critic. Faster improvements on learning curve at optimal optimization length (o= 8 versus 4 with s=24) means that the sampled TD offers better variance on the reduced bias by the scanned initial hidden state. In terms of behavior, the scanning strategy enables the agent to appropriately learn from experiences. This was not the case for the agent with the hidden state $h_{-1}$ initialized as zero (blue line in Fig.\ref{fig:scan}).

Another evidence of the bias reduction on TD by scanning is that the optimal update length changes from 4 to 8 when the optimization length is 8 (Fig.\ref{fig:optupdlen_easy}). Since the scanning method enables $h_{-1}$ to represent meaningful history of the optimization interval, TD's in early time-steps (Fig.\ref{fig:nstep_3}) in the interval properly contributes to training the Critic, hence the extension of $u$ upto $l$ improves the performance.

\subsubsection{Truncation of backpropagation through time} We found that the truncation of policy gradient through time improves the learning performance of our LSTM agent as shown in Fig. \ref{fig:bpttTrun} by 26.1 reward points. More importantly, the agent trained without BPTT behaves in desirable way that it negotiates upcoming obstacles at the right distance. For instance, the agent trained with BPTT kicks its leg too early before crossing gaps (Fn. \ref{fn:video}) ended up failure frequently on gaps, whereas the one without BPTT does not (Fig.\ref{fig:gap2}).

\subsubsection{Experience injection facilitates diverse behaviors}\label{sec:res-injection}
Interestingly, this study reveals that the RDPG agent can learn \textit{non-monotonic} motion behavior with the help of the experience injection. It is particularly useful in adapting to irregular terrains. Fig.\ref{fig:behavior} shows that the external experience injection takes advantage of the behaviors acquired by previous trained agents, as the new agent is capable of generating diverse set of agile behaviors to stably traverse over different terrain types. Our proposed use of experience injection also facilitates faster learning with the highest reward (Table \ref{table:summary-performance}). This is not the case for the DDPG agent. We suspect that the lack of memorization capability of past with Feedforward-network rather deters the correct policy gradient of the DDPG agent since the state and observation distributions are changed by our injection method.

As shown in Fig.\ref{fig:hurdle}, the agent trained with experience injection can learn how to jump over a hurdle and cross over by putting one leg ahead first (Fig.\ref{fig:hurdle_gap}). Previously, those two distinct behaviors were only achieved separately by different trained policies. These results show the agent trained with proposed experience injection is capable of combining the knowledge and behaviors from multiple policies, and hence can generate a better policy. Moreover, as shown in Fig.\ref{fig:hurdle} and \ref{fig:stair}, some novel responses with better optimality that naturally leads to agile motions emerge, such as stable landing as well as step-by-step climbing over stairs.

\section{Conclusion and future work}\label{sec:conclusion}
In this paper, we introduced several methods to improve induced bias and variance of error measure in principled way for RDPG algorithm in partially observable environment. Our work improved previous optimization methods in terms of error measurement, initial hidden state initialization, BPTT truncation and their correlation. Furthermore, we suggested an effective method of knowledge transfer for episodic experience replay given that the agent estimated better belief-state from observation via memory. The robot controlled by our proposed RDPG with this improved optimization process was trained in the OpenAI's simulation where the terrain and obstacles are partially observable and monotonic locomotion fails. Overall, our study provided evidence that the RDPG agent with proposed optimization methods improves policy via better belief-state representation. 


\section*{Acknowledgement}
This work is supported by the UK Robotics and Artificial Intelligence Hubs -- Offshore Robotics for Certification of Assets (ORCA), EP/R026173/1 funded by the Engineering and Physical Sciences Research Council (EPSRC), UK.





%

\bibliographystyle{IEEEtran}
\bibliography{root}

\end{document}